\documentclass[letterpaper]{article}
\usepackage{aaai}
\usepackage{times}
\usepackage{helvet}
\usepackage{courier}
\usepackage{url}
\usepackage{subfigure} 
\usepackage[mathscr]{eucal}
\usepackage{array}
\usepackage{multirow}
\usepackage{amsmath,amssymb}
\usepackage{tikz}
\usepackage{booktabs}
\usepackage{wrapfig}
\usepackage{caption} 

\newcommand\citep{\cite}
\newcommand\citet[1]{\citeauthor{#1}~\shortcite{#1}}

\def\plotheight{95pt}

\usetikzlibrary{arrows}
\tikzset{>=latex}

\newcommand\shrink[1]{}

\def\eql(#1,#2){{#1\!\!=\!#2}}
\def\miss{{\ensuremath{\mathsf{mi}}}}
\def\obs{{\ensuremath{\mathsf{ob}}}}
\def\unobs{{\ensuremath{\mathsf{unob}}}}

\def\data{{\boldsymbol{\mathscr{D}}}}

\def\mX{{\bf X}_\mathit{m}}

\def\oX{{\bf X}_\mathit{o}}
\def\R{{\bf R}}
\def\r{{\bf r}}

\def\bn{{\mathcal{N}}}

\def\n(#1){\bar{#1}}

\def\u{{\bf u}}
\def\U{{\bf U}}

\def\x{{\bf x}}
\def\X{{\bf X}}

\def\Y{{\bf Y}}
\def\W{{\bf W}}

\def\Z{{\bf Z}}

\def\pr{{\it Pr}}

\renewcommand{\Pr}{\operatorname{Pr}}

\DeclareMathOperator{\indep}{\perp\!\!\!\perp}

\nocopyright
\begin{document}
\title{Efficient Algorithms for Bayesian Network \\ Parameter Learning from Incomplete Data}
\author{
Guy Van den Broeck%
\thanks{Both authors contributed equally to this work. GVdB is also affiliated with KU Leuven, Belgium.}
{\rm \ and} 
Karthika Mohan\footnotemark[1]
{\rm \ and} 
Arthur Choi {\rm and} 
Judea Pearl \\
University of California, Los Angeles \\
Los Angeles, CA 90095 \\
\texttt{\{guyvdb,karthika,aychoi,judea\}@cs.ucla.edu}
}
\maketitle
\begin{abstract}
\begin{quote}
We propose an efficient family of algorithms to learn the parameters of a Bayesian network from incomplete data.  
In contrast to textbook approaches such as EM and the gradient method, our approach is non-iterative, yields closed form parameter estimates, and eliminates the need for inference in a Bayesian network. Our approach provides consistent parameter estimates for missing data problems that are MCAR, MAR, and in some cases, MNAR. Empirically, our approach is orders of magnitude faster than EM (as our approach requires no inference). Given sufficient data, we learn parameters that can be orders of magnitude more accurate.
\end{quote}
\end{abstract}

\section{Introduction}

When learning the parameters of a Bayesian network from data with missing values, the conventional wisdom among machine learning practitioners is that there are two options: either use \emph{expectation maximization}~(EM) or use likelihood \emph{optimization} with a gradient method; see, e.g., \citep{Darwiche09,KollerFriedman,MurphyBook,BarberBook}. 
These two approaches are known to consistently estimate the parameters when values in the data are \emph{missing at random} (MAR).
However, these two standard approaches suffer from the following disadvantages.  First, they are \emph{iterative}, and hence they may require many passes over a potentially large dataset.  Next, they \emph{require inference} in the Bayesian network, which is by itself already intractable (for high-treewidth networks with little local structure \citep{Chavira.Darwiche.Sat.2006,Chavira.Darwiche.Ijcai.2007}).  Finally, these algorithms may get stuck in \emph{local optima}, which means that, in practice, one must run these algorithms multiple times with different initial seeds, and keep those parameter estimates that obtained the best likelihood.

Recently, \citet{mohan2013missing} showed that the joint distribution of a Bayesian network can be recovered consistently from incomplete data, for all MCAR and MAR problems as well as a major subset of MNAR problems, when given access to a missingness graph. This graph is a formal representation of the \emph{causal mechanisms} responsible for missingness in an incomplete dataset. Using this representation, they are able to decide whether there exists a consistent estimator for a given query $Q$ (e.g., a joint or conditional distribution).  If the answer is affirmative, they identify a \emph{closed-form} expression to estimate $Q$ in terms of the observed data, which is asymptotically consistent. 

Based on this framework, we contribute a new and \emph{practical family of parameter learning algorithms} for Bayesian networks.
The key insight of our work is the following.
There exists a most-general, least-committed missingness graph that captures the MCAR or MAR assumption, but invokes no additional independencies.
Although this is a minor technical observation, it has far-reaching consequences.
It enables the techniques of \citeauthor{mohan2013missing} to be applied directly to MCAR or MAR data, without requiring the user to provide a more specific missingness graph. Hence, it enables our new algorithms to serve as drop-in replacements for the already influential EM algorithm in existing applications.
It results in practical algorithms for learning the parameters of a Bayesian network from an incomplete dataset that have the following advantages:
\begin{enumerate}
  \item the parameter estimates are efficiently computable in \emph{closed-form}, requiring only a \emph{single pass over the data}, as if no data was missing,
  \item the parameter estimates are obtained, \emph{inference-free}, in the Bayesian network, and
  \item the parameter estimates are \emph{consistent} when the values of a dataset are MCAR or MAR, i.e., we recover the true parameters as the dataset size approaches infinity.
\end{enumerate}
Advantages (1) and (2) are significant computational advantages over EM, in particular, when the dataset size is very large (cf., the Big Data paradigm), or for Bayesian networks that are intractable for exact inference.
Moreover, because of advantage (1), we do not use iterative optimization, and our estimates do not suffer from local optima.  Note further that all these advantages are already available to us when learning Bayesian networks from \emph{complete} datasets, properties which certainly contributed to the popularity of Bayesian networks today, as probabilistic models.

As secondary contributions, we show how to factorize estimates to extract more information from the data, and how to use additional information about the missingness mechanism to improve the convergence of our algorithms. Moreover, we present an initial experimental evaluation of the proposed algorithms, illustrating their key properties.

\shrink{
The second set of algorithms, that we propose, assumes that the structure of a missingness graph is known, perhaps from a domain expert. 
If the graph agrees with the MAR assumption, our first set of algorithms can be used. Nevertheless, we show that using the more specific missingness graph improves the convergence of our estimates.
If the graph does not agree with the MAR assumption, EM fails to recover the true parameters while our method can still estimate them consistently by exploiting the conditional independencies encoded in the graph.
}

\section{Technical Preliminaries} \label{sec:prelim}

In this paper, we use upper case letters (\(X\)) to denote variables
and lower case letters \((x)\) to denote their values.  Variable sets
are denoted by bold-face upper case letters (\(\X\)) and their
instantiations by bold-face lower case letters (\(\x\)).  
Generally, we will use \(X\) to denote a variable in a Bayesian
network and \(\U\) to denote its parents. A network parameter will
therefore have the general form \(\theta_{x|\u}\), representing the
probability \(\pr(\eql(X,x)|\eql(\U,\u))\).

\begin{figure}[tb]
  \centering
  \subfigure[Dataset \(\data\) and DAG]{%
  \begin{minipage}{.40\linewidth}
    \centering
    \begin{minipage}{0.3\columnwidth}
      \footnotesize
      \centering
      \begin{tabular}{cc}
        \(X\)     & \(Y\)     \\\hline
        \(\n(x)\) & \(y\)     \\
        \(x\)     & ?         \\
        \(x\)     & \(\n(y)\) \\
        \(\n(x)\) & ?         \\
        \multicolumn{2}{c}{\dots}
      \end{tabular}
    \end{minipage}~
    \begin{minipage}{0.6\columnwidth}
      \centering
      \includegraphics[scale=0.4,clip=true,angle=0]{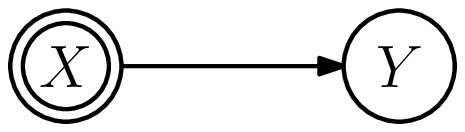}
    \end{minipage}
  \end{minipage}
    \label{fig:example:a}
  }
  \subfigure[Missingness Dataset \(\data\) and DAG]{%
\footnotesize
\begin{minipage}{0.56\linewidth}
  \centering
  \begin{minipage}{0.45\columnwidth}
    \centering
    \begin{tabular}{cc|cc}
      \(X\)     & \(Y\) & \(Y^\star\) & \(R_Y\) \\\hline
      \(\n(x)\) & \(y\)     & \(y\)       & \obs \\
      \(x\)     & ?     & \miss       & \unobs \\
      \(x\)     & \(\n(y)\)     & \(\n(y)\)   & \obs \\
      \(\n(x)\) & ?     & \miss       & \unobs \\
      \multicolumn{2}{c}{\dots} & \multicolumn{2}{c}{\dots}
    \end{tabular}
  \end{minipage}~
  \begin{minipage}{0.48\columnwidth}
    \centering
    \includegraphics[scale=0.4,clip=true,angle=0]{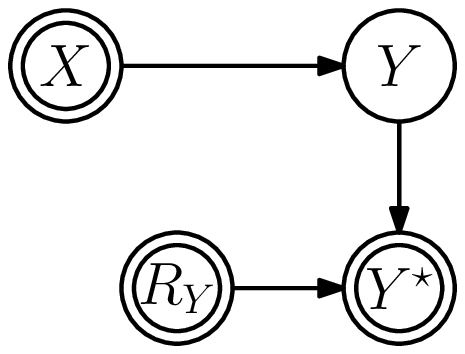}
  \end{minipage}
\end{minipage}
    \label{fig:example:b}
  }
  \caption{Datasets and DAGs.}
  \label{fig:example}
\end{figure}

As an illustrative example, consider Figure~\ref{fig:example:a},
depicting a dataset \(\data\), and the directed acyclic graph (DAG)
\(\mathcal{G}\) of a Bayesian network, both over variables \(X\) and
\(Y\).  Here, the value for variable \(X\) is always observed in the
data, while the value for variable \(Y\) can be missing.  In the
graph, we denote a variable that is always observed with a
double-circle.  Now, if we happen to know the mechanism that causes
the value of \(Y\) to become missing in the data, we can include it in
our model, as depicted in Figure~\ref{fig:example:b}.  Here, the
missingness of variable \(Y\), which is reported by variable
\(Y^\star\), depends on the value of another variable \(R_Y\).
This graph is called a \emph{missingness} graph, and 
can serve as a useful tool for analyzing missing data problems
\citep{mohan2013missing,Darwiche09,KollerFriedman}.


In our example, we have augmented the dataset and graph with new variables. Variable \(R_Y\) represents the causal mechanism that dictates the missingness of the value of \(Y\).  This mechanism can be active (\(Y\) is unobserved), which we denote by \(\eql(R_Y,\unobs)\).  Otherwise, the mechanism is passive (\(Y\) is observed), which we denote by \(\eql(R_Y,\obs)\).  Variable \(Y^\star\) acts as a proxy on the value of \(Y\) in the data, which may be an observed value \(y\), or a special value (\miss) when the value of \(Y\) is missing.  The value of \(Y^\star\) thus depends functionally on variables \(R_Y\) and \(Y\), with a corresponding CPT:
\begin{align*}
  Y^\star = f(R_{Y},Y)  &= \left\{
  \begin{tabular}{ll}
    \miss & if \(R_{Y}=\unobs\) \\
    \(Y\) & if \(R_{Y}=\obs\)
  \end{tabular}\right.\\
\Pr(Y^\star|Y,R_Y) &= \left\{
\begin{tabular}{rl}
1 & \mbox{if \(\eql(R_Y,\unobs)\) and \(Y^\star = \miss\)} \\
1 & \mbox{if \(\eql(R_Y,\obs)\) and \(Y^\star = Y\)} \\
0 & \mbox{otherwise.}
\end{tabular}
\right.
\end{align*}
That is, when \(\eql(R_Y,\unobs)\), then \(Y^\star = \miss\); otherwise \(\eql(R_Y,\obs)\) and the proxy \(Y^\star\) assumes the observed value of variable \(Y\).
Using missingness graphs, one can analyze the
distribution over the observed and missing values, and relate it to
the underlying distribution that we seek to estimate from data.  As
\citet{mohan2013missing} further show, one can exploit the conditional
independencies that these graphs encode, in order to extract
\emph{consistent} estimates from missing data problems, including MNAR
ones, whose underlying assumptions would put itself out of the scope
of existing techniques.

We now more formally define the missing data problems that we consider in this paper.  When learning a Bayesian network \(\bn\) from an incomplete dataset \(\data\), there is an underlying but unknown distribution \(\Pr(\X)\) that is induced by the network \(\bn\) that we want to learn.  The variables \(\X\) are partitioned into two sets: the fully-observed variables \(\X_o\), and the partially-observed variables \(\X_m\) that have missing values in the data.  We can take into account the mechanisms that cause the values of variables \(\X_m\) to go missing, as in our example above, by introducing variables \(\R\) representing the \emph{causal mechanisms} themselves, and variables \(\X_m^\star\) that act as \emph{proxies} to the variables \(\X_m\).  This augmented Bayesian network, which we refer to as the missingness graph \(\bn^\star\), now has variables \(\X_o,\X_m^\star,\R\) that are fully-observed, and variables \(\X_m\) that are only partially-observed.  Moreover, network \(\bn^\star\) induces a distribution \(\Pr(\X_o,\X_m,\X_m^\star,\R)\) which now embeds the original distribution \(\Pr(\X_o,\X_m)\) of network \(\bn\) as a marginal distribution.

Recently, \citet{mohan2013missing} identified conditions on the missingness graph \(\bn^\star\) 
that allow the original, partially-observed distribution
\(\Pr(\X_o,\X_m)\) to be identified from the fully-observed
distribution \(\Pr(\X_o,\X^\star_m,\R)\).  However, in practice, we
only have access to a dataset \(\data\), and the
corresponding \emph{data distribution} that it induces:
\[
\textstyle
\Pr_\data(\x_o,\x^\star_m,\r) = \frac{1}{N} \data\#(\x_o,\x^\star_m,\r),
\]
where \(N\) is the number of instances in dataset \(\data\), and where
\(\data\#(\x)\) is the number of instances where instantiation \(\x\)
appears in the data.\footnote{Note that the data distribution is
  well-defined over the variables \(\X_o,\X^\star_m\) and \(\R\), as
  they are fully-observed in the augmented dataset, and that \(\Pr_\data\) can be represented compactly in space linear in~\(N\), as we need not explicitly represent those
  instantiations \(\x\) that were not observed in the data.}
However, the data distribution \(\Pr_\data\) tends to the true
distribution \(\Pr\) (over the fully-observed variables), as \(N\) tends to infinity. 


In this paper, we show how we can leverage the results of \citet{mohan2013missing}, even when we do not have access to the complete missingness graph that specifies the direct causes, or parents, of the missingness mechanisms \(\R\). We identify practical and efficient algorithms for the consistent estimation of Bayesian network parameters.  First, we only assume general conditions that hold in broad classes of missingness graphs, which further characterize commonly-used assumptions on missing data.
Subsequently, we show how to exploit more specific knowledge of the underlying missingness graph that is available (say, from a domain expert), to obtain improved parameter estimates.

\paragraph{Missingness Categories} 
An incomplete dataset is categorized as \emph{Missing Completely At Random} (MCAR) if all mechanisms \(\R,\) that cause the values of variables \(\X_m\) to go missing, are marginally independent of $\X$, i.e., where \((\X_m,\X_o) \indep \R \). This corresponds to a missingness graph where no variable in $\X_m \cup \X_o$ is a parent of any variable in \(\R\). Note that the example graph that we started with in Section~\ref{sec:prelim} implies an MCAR dataset. 
        
An incomplete dataset is categorized as \emph{Missing At Random} (MAR) if missingness mechanisms are conditionally independent of the partially-observed variables given the fully-observed variables, i.e., if \(\X_m \indep \R \mid \X_o\). This corresponds to a missingness graph where variables \(\R\) are allowed to have parents, as long as none of them are partially-observed.  In the example missingness graph of Section~\ref{sec:prelim}, adding an edge \(X \to R_Y\) results in a graph that yields MAR data.  
This is a stronger, variable-level definition of MAR, which has previously been used in the machine learning literature \citep{Darwiche09,KollerFriedman}, in contrast to the event-level definition of MAR, that is prevalent in the statistics literature \citep{rubin1976inference}. 

An incomplete dataset is categorized as \emph{Missing Not At Random} (MNAR) if it is neither MCAR nor MAR.  In the example graph of Section~\ref{sec:prelim}, adding an edge \(Y \to R_Y\) yields an MNAR assumption.

\section{Closed-Form Learning Algorithms}

We now present a set of algorithms to learn the parameters \(\theta_{x|\u}\) of a Bayesian network $\bn$ from the data distribution $\Pr_{\data}$ over the fully-observed variables in the augmented dataset.
We do so for different missing data assumptions, but without knowing the missingness graph that generated the data. To estimate the conditional probabilities \(\theta_{x|\u}\) that parameterize a Bayesian network, we estimate the joint distributions \(\Pr(X,\U)\), which are subsequently normalized.  Hence, it suffices, for our discussion, to estimate marginal distributions \(\Pr(\Y)\), for families \(\Y = \{X\} \cup \U\).  Here, we let $\Y_o = \Y \cap \X_o$ denote the observed variables in \(\Y\), and $\Y_m = \Y \cap \X_m$ denote the partially-observed variables.  Further, we let $\R_\Z \subseteq \R$ denote the missingness mechanisms for a set of partially-observed variables~\(\Z\).
Appendix~\ref{app:example} illustrates our learning algorithms on a concrete dataset.

\subsection{Direct Deletion for MCAR} \label{s:d-mcar}

The statistical technique of \emph{listwise deletion} is perhaps the
simplest technique for performing estimation with MCAR data: we simply
delete all instances in the dataset that contain missing values, and
estimate our parameters from the remaining dataset, which is now
complete.  Of course, with this technique, we potentially ignore large
parts of the dataset.  The next simplest technique is perhaps pairwise
deletion, or available-case analysis: when estimating a quantity over
a pair of variables \(X\) and \(Y\), we delete just those instances
where variable \(X\) or variable \(Y\) is missing.

Consider now the following deletion technique, which is expressed in
the terms of causal missingness mechanisms, which we reviewed in the
previous section.  In particular, to estimate the marginals
\(\Pr(\Y)\) of a set of (family) variables \(\Y\), from the data
distribution \(\Pr_\data\), we can use the estimate:
\begin{align*}
\Pr(\Y)
& = \Pr(\Y_o,\Y_m | \eql(\R_{\Y_m},\obs))
  \tag*{by \(\oX,\mX \indep \R\)}\\
& = \Pr(\Y_o,\Y_m^\star | \eql(\R_{\Y_m},\obs))  \tag*{by \(\eql(\mX ,\mX^\star)\) when \(\eql(\R,\obs)\)}\\
  & \approx \Pr_\data(\Y_o,\Y_m^\star | \eql(\R_{\Y_m},\obs))
\end{align*}
That is, we can estimate \(\Pr(\Y)\) by simply using the subset of the data where every variable in \(\Y\) is observed (which follows from the assumptions implied by MCAR data).  Because the data distribution \(\Pr_\data\) tends to the true distribution \(\Pr\), this implies a consistent estimate for the marginals \(\Pr(\Y)\).  In contrast, the technique of listwise deletion corresponds to the estimate \(\Pr(\Y) \approx \Pr_\data(\Y_o,\Y_m^\star | \eql(\R_{\X_m},\obs))\), and the technique of pairwise deletion corresponds to the above, when \(\Y\) contains two variables.
To facilitate comparisons with more interesting estimation algorithms that we shall subsequently consider, we refer to the more general estimation approach above as \emph{direct deletion}.

\subsection{Direct Deletion for MAR} \label{s:d-mar}

In the case of MAR data, we cannot use the simple deletion techniques
that we just described for MCAR data---the resulting
estimates would not be consistent.
However, we show next that it is possible to obtain consistent
estimates from MAR data, using a technique that is as simple and
efficient as direct deletion.  Roughly, we can view this technique as
deleting certain instances from the dataset, but then re-weighting the
remaining ones, so that a consistent estimate is obtained.
This is the key contribution of this
paper, which provides a new algorithm with desirable properties
(compared to EM), as described in the introduction.  
We shall subsequently show how to obtain even better estimates, later.

Again, to estimate network parameters \(\theta_{x|\u},\) it suffices
to show how to estimate family marginals $\Pr(\Y),$ now under the MAR
assumption.  Let $\X_o' = \X_o \setminus \Y_o$ denote the
fully-observed variables outside of the family variables \(\Y\) (i.e.,
\(\X_o = \Y_o \cup \X'_o\)).  We have
\begin{align*}
\Pr(\Y)
& = \sum_{\X_o'} \Pr(\Y_o,\Y_m,\X_o') \\
&  = \sum_{\X_o'} \Pr(\Y_m|\Y_o,\X'_o) \Pr(\Y_o,\X'_o)
\end{align*}
Hence, we reduced the problem to estimating two sets of probabilities.
Estimating the probabilities \(\Pr(\Y_o,\X'_o)\) is straightforward, as variables \(\Y_o\) and \(\X'_o\) are fully observed in the data.
The conditional probabilities $\Pr(\Y_m|\Y_o,\X'_o)$ contain partially observed variables \(\Y_m\), but they are conditioned on all fully observed variables~$\X_o = \Y_o \cup \X'_o$. The MAR definition implies that each subset of the data that fixes a value for $\X_o$ is locally MCAR. 
Analogous to the MCAR case, we can estimate each conditional probability as
\begin{align*}
\Pr(\Y_m|\Y_o,\X'_o) = \Pr(\Y^\star_m|\Y_o,\X'_o,\eql(\R_{\Y_m},\obs)).
\end{align*}
This leads to the following algorithm,
\begin{align*}
\Pr(\Y)
\approx \sum_{\X_o'} \Pr_\data(\Y^\star_m|\Y_o,\X'_o,\eql(\R_{\Y_m},\obs)) \Pr_\data(\Y_o,\X'_o)
\end{align*}
which uses only the fully-observed variables of the data distribution $\Pr_\data$.\footnote{Note that the summation requires only a single pass through the data, i.e., for only those instantiations of \(\X_o'\) that appear in it.}
Again, \(\Pr_\data\) tends to the true distribution \(\Pr\), as the dataset size tends to infinity, implying a consistent estimate of \(\Pr(\Y).\) 

\subsection{Factored Deletion} 
 
We now propose a class of deletion algorithms that exploit more data than direct deletion.  In the first step, we generate multiple but consistent estimands for the query so that each estimand utilizes different parts of a dataset to estimate the query. In the second step, we aggregate these estimates to compute the final estimate and thus put to use almost all tuples in the dataset. Since this method exploits more data than direct deletion, it obtains a better estimate of the query.

\paragraph{Factored Deletion for MCAR } \label{s:f-mcar}

Let the query of interest be $\Pr(\Y)$, and let $ Y^1,Y^2,\dots,Y^{n}$ be any ordering of the $n$ variables in $\Y$.  Each ordering yields a unique factorization, i.e., 
 $\Pr(\Y)=\prod_{i=1}^n \Pr\left(Y^i \middle| Y^{i+1},\dots,Y^{n}\right)$. 
We can estimate each of these factors independently, on the subset of the data in which all of its variables are fully observed (as in direct deletion), i.e., $\Pr\left(Y^i \middle| Y^{i+1},\dots,Y_m^{n}\right)=\Pr\left(Y^i \middle| Y^{i+1},\dots,Y_m^{n}, \eql(\R_{Z^i}, \obs)\right)$ where $\Z^i$ is the set of partially observed variables in the factor. When $|\Y \cap \X_m| >1$, we can utilize much more data compared to direct deletion. We refer to Appendix~\ref{sec:fmcarexample} for an example. 
\begin{figure}[tb]
  \centering
\includegraphics[width=0.8\linewidth]{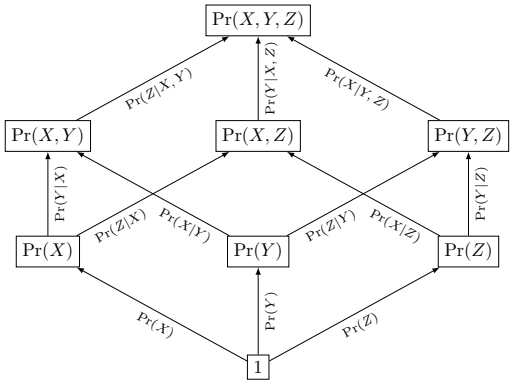}
  \caption{Factorization Lattice of $\Pr(X,Y,Z)$} \label{fig:lattice}
\end{figure}

So far, we have discussed how a consistent estimate of $\Pr(\Y)$ may be computed given a factorization. Now we shall detail how estimates from each factorization can be aggregated to compute more accurate estimates of $\Pr(\Y)$.  Let \(k\) be the number of variables in a family \(\Y\).  The number of possible factorizations is \(k!\).  However, different factorizations share the same sub-factors, which we can estimate once, and reuse across factorizations.  We can organize these computations using a lattice, as in Figure~\ref{fig:lattice}, which has only $2^k$ nodes and $k2^{k-1}$ edges. Our algorithm will compute as many estimates as there are edges in this lattice, which is only on the order of \(O(n \log n)\), where \(n\) is the number of parameters being estimated for a family \(Y\) (which is also exponential in the number of variables \(k\)).

More specifically, our factored deletion algorithm works as follows.
First, we estimate the conditional probabilities on the edges of the lattice, each estimate using the subset of the data where its variables are observed.
Second, we propagate the estimates, bottom-up. 
For each node, there may be several alternative estimates available, on its incoming edges. There are various ways of aggregating these estimates, such as mean, median, and propagating the lowest-variance estimate.\footnote{We use an inverse-variance weighting heuristic, which was somewhat better in our experiments.}
Whereas direct deletion uses only those instances in the data where \emph{all} variables in $\Y$ are observed, factored deletion uses any instance in the data where \emph{at least one} variable in $\Y$ is observed.


\paragraph{Factored Deletion for MAR} \label{s:f-mar}
Let $Y_m^1,Y_m^2,\dots,Y_m^{n}$ be any ordering of the $n$ partially observed variables $\Y_m \subseteq \Y$ and let $\X_o' = \X_o \setminus \Y_o$ denote the
fully-observed variables outside of \(\Y\). Given an ordering we can factorize $\Pr(\Y)$ as $\sum_{\X_o^\prime} \Pr(\Y_o,\X_o^\prime) \prod_{i=1}^n \Pr\left(Y_m^i \middle| \Z_m^{i+1},\X_o\right)$
 where \(\Z_m^i=\left\{Y_m^j \middle| i \leq j \leq n\right\}\).  We then proceed in a manner similar to factored deletion for MCAR to estimate individual factors and aggregate estimates to compute $\Pr(\Y)$. For equations and derivations, please see Appendix~\ref{sec:fmarextra}.

\subsection{Learning with a Missingness Graph}

We have so far made very general assumptions about the structure of the missingness graph, capturing the MCAR and MAR assumptions.  
In this section, we show how to exploit additional
knowledge about the missingness graph to further improve the quality of our estimates.

\subsubsection*{Informed Deletion for MAR}
Suppose now that we have more in-depth knowledge of the missing data mechanisms
of an MAR problem, namely that we know the subset $\W_o$ of the observed variables 
$\X_o$ that suffice to separate the missing values from their causal mechanisms, i.e., where \(\X_m \indep \R \mid \W_o\).
We can exploit such knowledge in our direct deletion algorithm, to obtain improved parameter estimates.  In particular, we can reduce the scope of the summation
in our direct deletion algorithm from the variables \(\X'_o\) (the set of variables in \(\X_o\) that lie outside the family \(\Y\)), to the variables \(\W_o'\) (the set of variables in
\(\W_o\) that lie outside the family \(\Y\)),\footnote{Again, we need only consider, in the summation, the instantiations of \(\W_o'\) that appear in the dataset.} yielding the algorithm:
\begin{align*}
\Pr(\Y)
\approx \sum_{\W_o'} \Pr_\data(\Y^\star_m|\Y_o,\W'_o,\eql(\R_{\Y_m},\obs)) \Pr_\data(\Y_o,\W'_o)
\end{align*}
We refer to this algorithm as \emph{informed direct deletion}.
By reducing the scope of the summation, we need to estimate fewer sub-expressions $\Pr_\data(\Y^\star_m|\Y_o,\W'_o,\eql(\R_{\Y_m},\obs))$. 
This results in a more efficient computation, but further, each individual sub-expression can be estimated on more data.
Moreover, our estimates remain consistent.
We similarly replace $\X_o$ by $\W_o$ in the factored deletion algorithm, yielding the \emph{informed factored deletion} algorithm. Appendix~\ref{sec:mnar} presents an empirical evaluation of informed deletion and exemplifies cases where knowing the missingness graph lets us consistently learn from MNAR data, which is beyond the capabilities of maximum-likelihood learners.

\section{Empirical Evaluation} \label{sec:experiments}

To evaluate the proposed learning algorithms, we simulate partially observed datasets from Bayesian networks, and re-learn their parameters from the data.\footnote{The implementation and experimental setup is available at \url{http://reasoning.cs.ucla.edu/deletion}}
We consider the following algorithms:
\begin{description}
  \item[D-MCAR \& F-MCAR] direct/factored deletion for MCAR data.
  \item[D-MAR \& F-MAR] direct/factored deletion for MAR data.
  \item[EM-$k$-JT] EM with $k$ random restarts and using the jointree inference algorithm.
  \item[F-MAR \texttt{+} EM-JT] EM using the jointree inference algorithm, seeded by the F-MAR estimates.
\end{description}
Remember that D-MCAR and F-MCAR are consistent for MCAR data only, while D-MAR and F-MAR are consistent for general MAR data. EM is consistent for MAR data, but only if it converges to maximum-likelihood estimates.

\shrink{
Given the multitude of Bayesian networks in use today, and the different types of missing data mechanisms one can conceive of, it is not feasible to provide a conclusive practical comparison of these algorithms for all learning problems.
Instead, we choose to report on individual experiments that we find representative of general trends, and that highlight trade-offs between the learning algorithms. 
For a more exhaustive comparison, with six different Bayesian networks and many different missing data mechanisms, we refer the reader to the supplementary online material.
}

We evaluate the learned parameters in terms of their
\emph{likelihood} on independently generated, fully-observed test data, and the \emph{Kullback-Leibler divergence}~(KLD) between the original and learned Bayesian networks.
We report per-instance log-likelihoods (which are divided by dataset size).
We evaluate the learned models on unseen data, so all learning algorithms assume a symmetric Dirichlet prior on the network parameters with a concentration parameter of $2$.




Appendix~\ref{sec:exp:mcar} provides empirical results on the simpler case of learning from \emph{MCAR data}, where all algorithms are consistent. As expected, all produce increasingly accurate estimates as more data becomes available.  Compared to EM, where the complexity of inference prevents scaling to large datasets, D-MCAR and F-MCAR can obtain more accurate estimates orders-of-magnitude faster.

In this section, we investigate the more challenging problem of learning from \emph{MAR data}, which are generated as follows:
  (a) select an $m$-fraction of the variables to be partially observed,
  (b) introduce a missingness mechanism variable $R_X$ for each partially observed variable $X$,
  (c) assign $p$ parents to each $R_X$, that are randomly selected from the set of observed variables, giving preference to neighbors of $X$ in the network,
  (d) sample parameters for the missingness mechanism CPTs from a Beta distribution,
  (e) sample a complete dataset with $R_X$ values, and
  (f) hide $X$ values accordingly.


\begin{figure*}[ht]
  \centering
  \subfigure[KL Divergence - Fire Alarm \label{fig:mar:small}]{%
    \includegraphics[height=\plotheight]{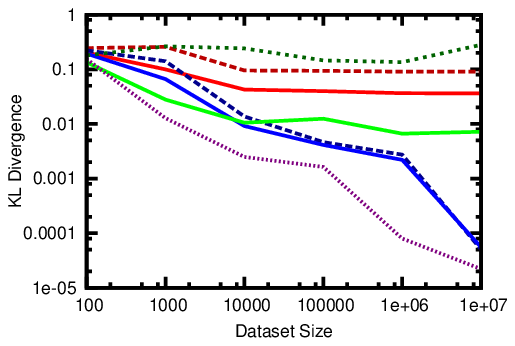}
  }
  \subfigure[Likelihood vs.\ Size - Alarm \label{fig:mar:big:size}]{%
    \includegraphics[height=\plotheight]{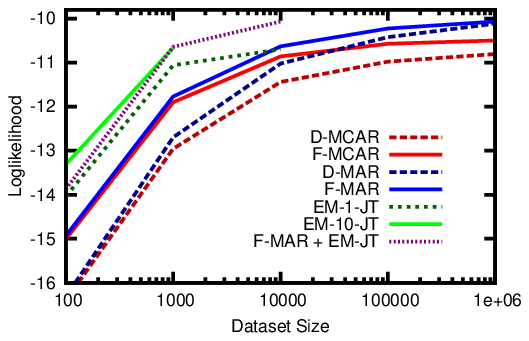}
  }
  \subfigure[Likelihood vs.\ Time - Alarm \label{fig:mar:big:time}]{%
    \includegraphics[height=\plotheight]{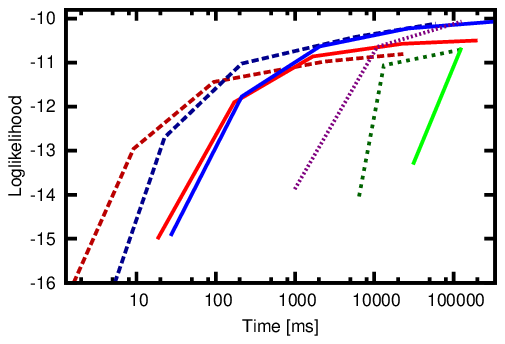}
  }
  \caption{Learning small, tractable Bayesian networks from MAR data (legend in \subref{fig:mar:big:size}).}
\end{figure*}

For our first MAR experiment, we work with a small network that is tractable enough for EM to scale to large dataset sizes.
Figure~\ref{fig:mar:small} shows KLD for the ``Fire Alarm'' network, which has only 6 variables (and hence, the complexity of inference is negligible). The missing data mechanisms were generated with $m=0.3$, $p=2$, and a Beta distribution with shape parameters $1.0$ and $0.5$.
All numbers are averaged over 64 repetitions with different random learning problems.

There is a significant difference between EM, with and without restarts, indicating that the likelihood landscape is challenging to optimize (compared to MCAR in Appendix~\ref{sec:exp:mcar}). EM-10-JT performs well for small dataset sizes, but stops converging after around 1,000 instances. This could be due to all restarts getting stuck in local optima.
The KLD of F-MAR starts off between EM-1-JT and EM-10-JT for small sizes, but quickly outperforms EM. For the largest dataset sizes, it learns networks whose KLD is several orders of magnitude smaller than EM.
The KLD improves further when we use F-MAR estimates to seed EM, although EM will not scale to larger, intractable networks.
D-MCAR and F-MCAR are not consistent for MAR data, and indeed converge to a biased estimate with a KLD around 0.1.
The factorized algorithms outperform their direct counterparts.

For our second MAR experiment, we work with the classical ``Alarm'' network, which has 37 variables. The missing data mechanisms were generated with $m=0.9$, $p=2$, and a Beta distribution with shape parameters $0.5$.
All reported numbers are averaged over 32 repetitions, and when no number is reported, a 10 minute time limit was exceeded.

Figures~\ref{fig:mar:big:size} and~\ref{fig:mar:big:time} show test set likelihood as a function of dataset \emph{size} and learning \emph{time}.
EM-10-JT performs well for very small dataset sizes, and again outperforms EM-1-JT. However, inference time is non-negligible and EM-10-JT fails to scale beyond 1,000 instances, whereas EM-1-JT scales to 10,000. 
The closed-form learners dominate all versions of EM as a function of time, and scale to dataset sizes that are two orders of magnitude larger.
EM seeded by F-MAR achieves similar quality to EM-10-JT, while being significantly faster than other EM learners.

\begin{table*}[htb]
\scriptsize
  \centering
  \caption{Log-likelihoods of large, intractable networks learned from MAR data (25 min.~time limit).}
  \label{table:bp}
  \begin{tabular}{@{}r!{\vrule width 1pt}r|rr|rr|rr!{\vrule width 1pt}r|rr|rr|rr@{}} \toprule
Size &  & EM-JT & EM-BP & D-MCAR & F-MCAR & D-MAR & F-MAR &  & EM-JT & EM-BP & D-MCAR & F-MCAR & D-MAR & F-MAR\\
\hline
$10^{2}$ & \multirow{5}{*}{\rotatebox[origin=c]{90}{Grid 90-20-1}} & - & \textbf{-49.15} & -80.00 & -56.45 & -79.81 & -55.94 & \multirow{5}{*}{\rotatebox[origin=c]{90}{Water}} & -18.88 & \textbf{-18.73} & -25.84 & -22.11 & -25.87 & -22.25 \\
$10^{3}$ & & - & -53.64 & -38.14 & -29.32 & -37.75 & \textbf{-29.09} & & -17.63 & \textbf{-14.41} & -18.39 & -15.95 & -18.27 & -15.79 \\
$10^{4}$ & & - & -85.65 & -26.21 & -23.05 & -25.45 & \textbf{-22.62} & & - & -14.52 & -15.57 & -14.07 & -15.24 & \textbf{-13.92} \\
$10^{5}$ & & - & - & -22.78 & -21.54 & -21.60 & \textbf{-20.79} & & - & -24.99 & -14.17 & -13.46 & -13.71 & \textbf{-13.19} \\
$10^{6}$ & & - & - & - & - & - & - & & - & - & \textbf{-13.73} & - & - & - \\
\hline
$10^{2}$ & \multirow{4}{*}{\rotatebox[origin=c]{90}{Munin 1}} & - & \textbf{-99.15} & -114.76 & -106.07 & -114.66 & -105.12 & \multirow{4}{*}{\rotatebox[origin=c]{90}{Barley}} & -89.05 & -89.15 & -89.57 & -89.17 & -89.62 & \textbf{-89.03} \\
$10^{3}$ & & - & -67.85 & -74.18 & -67.81 & -73.82 & \textbf{-67.39} & & - & -70.38 & -71.86 & -70.54 & -71.87 & \textbf{-70.27} \\
$10^{4}$ & & - & -66.62 & -57.50 & -54.94 & -56.96 & \textbf{-54.64} & & - & -76.48 & -56.37 & \textbf{-55.13} & -56.33 & - \\
$10^{5}$ & & - & - & -53.07 & \textbf{-51.66} & -52.27 & - & & - & - & -51.31 & - & \textbf{-51.19} & - \\
\bottomrule
\end{tabular}

\end{table*} 

For our third MAR experiment, Table~\ref{table:bp} reports results on four larger networks where exact inference is challenging. Each method is given a time
limit of 25 minutes, and data is generated as above. Appendix~\ref{sec:extended} provides further results, on more settings. We consider the following algorithms.
\begin{description}
  \item[EM-JT] The EM-10-JT algorithm used in anytime fashion, which returns, given a time limit, the best parameters found so far in any restart, even if EM did not converge. 
  \item[EM-BP] A variant of EM-JT that uses (loopy) belief propagation for (approximate) inference (in the E-step).
\end{description} 
We see that EM-JT, which performs exact inference, does not scale well to these networks. This problem is mitigated by EM-BP, which performs  \emph{approximate} inference, yet we find that it also has difficulties scaling (dashed entries indicate that EM-JT and EM-BP did not finish 1 iteration of EM).  
In contrast, F-MAR, and particularly D-MAR, can scale to much larger datasets.  
As for accuracy, the F-MAR method typically obtains the best likelihoods (in bold) for larger datasets, although EM-BP can perform better on small datasets.  We further evaluated D-MCAR and F-MCAR, although they are not in general consistent for MAR data, and find that they scale even further, and can also produce relatively good estimates (in terms of likelihood).

\section{Conclusions and Related Work}

For estimating parameters in Bayesian networks, maximum likelihood (ML) estimation is the typical approach used, where for incomplete data, the common wisdom among machine learning practitioners is that one needs to use Expectation-Maximization (EM) or gradient methods~\citep{Dempster:EM,Lauritzen:95} (see also, e.g., \citet{Darwiche09,KollerFriedman,MurphyBook,BarberBook}).  As we discussed, such methods do not scale well to large datasets or complex Bayesian networks as they (1) are iterative, (2) require inference in a Bayesian network, and (3) suffer from local optima.  
Considerable effort has been expended in improving on EM, across these dimensions, in order to, for example, (1) accelerate the convergence of EM, and to intelligently sample subsets of a dataset, e.g., \citet{ThiessonMH01}, (2) use approximate inference algorithms in lieu of exact ones when inference is intractable, e.g., \citet{GhahramaniJ97,MCEM}, and (3) escape local optima, e.g., \citet{ElidanNFS02}.  While EM is suitable for data that is MAR (the typical assumption, in practice), there are some 
exceptions, such as recent work on recommender systems that explicitly incorporate missing data mechanisms~\citep{marlin2009collaborative,reco,marlin2012collaborative}.

In the case of complete data, the parameter estimation task simplifies considerably, in the case of Bayesian networks: maximum likelihood estimates can be obtained inference-free and in closed-form, using just a single pass over the data: \(\theta_{x|\u} = \pr_\data(x|\u)\). 
In fact, the estimation algorithms that we proposed in this paper also obtain the same parameter estimates in the case of complete data, although we are not concerned with maximum likelihood estimation here---we simply want to obtain estimates that are consistent (as in estimation by the method of moments). 

In summary, we proposed an inference-free, closed-form method for consistently learning Bayesian network parameters, from MCAR and MAR datasets (and sometimes MNAR datasets, as in Appendix~\ref{sec:mnar}).
Empirically, we demonstrate the practicality of our method, showing that it is orders-of-magnitude more efficient than EM, allowing it to scale to much larger datasets.  Further, given access to enough data, we show that our method can learn much more accurate Bayesian networks as well.

Other inference-free estimators have been proposed for other classes of probabilistic graphical models.  For example, \citet{AbbeelKN06} identified a method for closed-form, inference-free parameter estimation in factor graphs of bounded degree from complete data.  More recently, \citet{HalpernS13} proposed an efficient, inference-free method for consistently estimating the parameters of noisy-or networks with latent variables, under certain structural assumptions.  We note that inference-free learning of the parameters of: Bayesian networks under MAR data (this paper), factor graphs of bounded degree, under complete data \citep{AbbeelKN06}, and structured noisy-or Bayesian networks with latent variables \citep{HalpernS13}, are all surprising results.  From the perspective of maximum likelihood learning, where evaluating the likelihood (requiring inference) seems to be unavoidable, the ability to consistently estimate parameters without the need for inference, greatly extends the accessibility and potential of such models.  
For example, it opens the door to practical structure learning algorithms, under incomplete data, which is a notoriously difficult problem in practice \citep{AbbeelKN06,JerniteHS13}.  

\bibliographystyle{aaai}
\bibliography{references}

\begin{thebibliography}{}

\bibitem[\protect\citeauthoryear{Abbeel, Koller, and Ng}{2006}]{AbbeelKN06}
Abbeel, P.; Koller, D.; and Ng, A.~Y.
\newblock 2006.
\newblock Learning factor graphs in polynomial time and sample complexity.
\newblock {\em Journal of Machine Learning Research} 7:1743--1788.

\bibitem[\protect\citeauthoryear{Barber}{2012}]{BarberBook}
Barber, D.
\newblock 2012.
\newblock {\em Bayesian Reasoning and Machine Learning}.
\newblock Cambridge University Press.

\bibitem[\protect\citeauthoryear{Caffo, Jank, and Jones}{2005}]{MCEM}
Caffo, B.~S.; Jank, W.; and Jones, G.~L.
\newblock 2005.
\newblock Ascent-based monte carlo expectation-maximization.
\newblock {\em Journal of the Royal Statistical Society. Series B (Statistical
  Methodology)} 67(2):pp. 235--251.

\bibitem[\protect\citeauthoryear{Chavira and
  Darwiche}{2006}]{Chavira.Darwiche.Sat.2006}
Chavira, M., and Darwiche, A.
\newblock 2006.
\newblock Encoding {CNFs} to empower component analysis.
\newblock In {\em Proceedings of SAT},  61--74.

\bibitem[\protect\citeauthoryear{Chavira and
  Darwiche}{2007}]{Chavira.Darwiche.Ijcai.2007}
Chavira, M., and Darwiche, A.
\newblock 2007.
\newblock Compiling {B}ayesian networks using variable elimination.
\newblock In {\em Proceedings of IJCAI},  2443--2449.

\bibitem[\protect\citeauthoryear{Darwiche}{2009}]{Darwiche09}
Darwiche, A.
\newblock 2009.
\newblock {\em Modeling and Reasoning with {B}ayesian Networks}.
\newblock Cambridge University Press.

\bibitem[\protect\citeauthoryear{Dempster, Laird, and
  Rubin}{1977}]{Dempster:EM}
Dempster, A.; Laird, N.; and Rubin, D.
\newblock 1977.
\newblock Maximum likelihood from incomplete data via the {EM} algorithm.
\newblock {\em Journal of the Royal Statistical Society {B}} 39:1--38.

\bibitem[\protect\citeauthoryear{Elidan \bgroup et al\mbox.\egroup
  }{2002}]{ElidanNFS02}
Elidan, G.; Ninio, M.; Friedman, N.; and Shuurmans, D.
\newblock 2002.
\newblock Data perturbation for escaping local maxima in learning.
\newblock In {\em Proceedings of AAAI},  132--139.

\bibitem[\protect\citeauthoryear{Ghahramani and Jordan}{1997}]{GhahramaniJ97}
Ghahramani, Z., and Jordan, M.~I.
\newblock 1997.
\newblock Factorial hidden markov models.
\newblock {\em Machine Learning} 29(2-3):245--273.

\bibitem[\protect\citeauthoryear{Halpern and Sontag}{2013}]{HalpernS13}
Halpern, Y., and Sontag, D.
\newblock 2013.
\newblock Unsupervised learning of noisy-or {B}ayesian networks.
\newblock In {\em Proceedings of UAI}.

\bibitem[\protect\citeauthoryear{Jernite, Halpern, and
  Sontag}{2013}]{JerniteHS13}
Jernite, Y.; Halpern, Y.; and Sontag, D.
\newblock 2013.
\newblock Discovering hidden variables in noisy-or networks using quartet
  tests.
\newblock In {\em Proceedings of NIPS},  2355--2363.

\bibitem[\protect\citeauthoryear{Koller and Friedman}{2009}]{KollerFriedman}
Koller, D., and Friedman, N.
\newblock 2009.
\newblock {\em Probabilistic Graphical Models: Principles and Techniques}.
\newblock MIT Press.

\bibitem[\protect\citeauthoryear{Lauritzen}{1995}]{Lauritzen:95}
Lauritzen, S.
\newblock 1995.
\newblock The {EM} algorithm for graphical association models with missing
  data.
\newblock {\em Computational Statistics and Data Analysis} 19:191--201.

\bibitem[\protect\citeauthoryear{Marlin and
  Zemel}{2009}]{marlin2009collaborative}
Marlin, B., and Zemel, R.
\newblock 2009.
\newblock Collaborative prediction and ranking with non-random missing data.
\newblock In {\em Proceedings of the third ACM conference on Recommender
  systems},  5--12.
\newblock ACM.

\bibitem[\protect\citeauthoryear{Marlin \bgroup et al\mbox.\egroup
  }{2007}]{marlin2012collaborative}
Marlin, B.; Zemel, R.; Roweis, S.; and Slaney, M.
\newblock 2007.
\newblock Collaborative filtering and the missing at random assumption.
\newblock In {\em Proceedings of UAI}.

\bibitem[\protect\citeauthoryear{Marlin \bgroup et al\mbox.\egroup
  }{2011}]{reco}
Marlin, B.; Zemel, R.; Roweis, S.; and Slaney, M.
\newblock 2011.
\newblock Recommender systems: missing data and statistical model estimation.
\newblock In {\em Proceedings of IJCAI}.

\bibitem[\protect\citeauthoryear{Mohan, Pearl, and
  Tian}{2013}]{mohan2013missing}
Mohan, K.; Pearl, J.; and Tian, J.
\newblock 2013.
\newblock Graphical models for inference with missing data.
\newblock In {\em Proceedings of NIPS}.

\bibitem[\protect\citeauthoryear{Murphy}{2012}]{MurphyBook}
Murphy, K.~P.
\newblock 2012.
\newblock {\em Machine Learning: A Probabilistic Perspective}.
\newblock MIT Press.

\bibitem[\protect\citeauthoryear{Pearl}{1987}]{pearl1987evidential}
Pearl, J.
\newblock 1987.
\newblock Evidential reasoning using stochastic simulation of causal models.
\newblock {\em Artificial Intelligence} 32(2):245--257.

\bibitem[\protect\citeauthoryear{Rubin}{1976}]{rubin1976inference}
Rubin, D.~B.
\newblock 1976.
\newblock Inference and missing data.
\newblock {\em Biometrika} 63(3):581--592.

\bibitem[\protect\citeauthoryear{Thiesson, Meek, and
  Heckerman}{2001}]{ThiessonMH01}
Thiesson, B.; Meek, C.; and Heckerman, D.
\newblock 2001.
\newblock Accelerating {EM} for large databases.
\newblock {\em Machine Learning} 45(3):279--299.

\bibitem[\protect\citeauthoryear{Tsamardinos \bgroup et al\mbox.\egroup
  }{2003}]{tsamardinos2003algorithms}
Tsamardinos, I.; Aliferis, C.~F.; Statnikov, A.~R.; and Statnikov, E.
\newblock 2003.
\newblock Algorithms for large scale {M}arkov blanket discovery.
\newblock In {\em Proceedings of FLAIRS}, volume 2003,  376--381.

\bibitem[\protect\citeauthoryear{Yaramakala and
  Margaritis}{2005}]{yaramakala2005speculative}
Yaramakala, S., and Margaritis, D.
\newblock 2005.
\newblock Speculative markov blanket discovery for optimal feature selection.
\newblock In {\em Proceedings of ICDM}.

\end{thebibliography}

%
\appendix

\section{Factored deletion for MAR}
\label{sec:fmarextra}

We now give a more detailed derivation of the factored deletion algorithm for MAR data.
Let the query of interest be $\Pr(\Y)$, and let $\X_o'=\X_m \setminus \Y_m$ and \(\Z_m^i=\left\{Y_m^j \middle| i \leq j \leq n\right\}\). 
We can then factorize the estimation of $\Pr(\Y)$ as follows.
\begin{align*}
\Pr(\Y) & = \sum_{\X_o^\prime} \Pr(\Y_m,\Y_o,\X_o^\prime) \\
& = \sum_{\X_o^\prime} \Pr(\Y_o,\X_o^\prime) \Pr(\Y_m|\Y_o,\X_o^\prime) \\
& = \sum_{\X_o^\prime} \Pr(\X_o) \Pr(\Y_m|\X_o) \\
& = \sum_{\X_o^\prime} \Pr(\X_o) \prod_{i=1}^n \Pr\left(Y_m^i \middle| \Z_m^{i+1},\X_o\right) \\
& = \sum_{\X_o^\prime} \Pr(\X_o) \prod_{i=1}^n \Pr\left(Y_m^i \middle| \Z_m^{i+1},\X_o,\eql(\R_{\Z_m^{i}},\obs)\right)
\end{align*}
The last step makes use of the MAR assumption.
This leads us to the following algorithm, based on the data distribution $\Pr_\data$, and the fully-observed proxy variables $Y_m^{i,\star}$ and $\Z_m^{i+1,\star}$.
\begin{align*}
\Pr(\Y) \!\approx\! \sum_{\X_o^\prime}  \Pr_\data\!(\X_o) \! \prod_{i=1}^n \Pr_\data\!\left(Y_m^{i,\star} \middle| \Z_m^{i+1,\star},\X_o, \eql(\R_{\Z_m^{i}},\obs)\right)
\end{align*}

\section{Learning with a Missingness Graph} \label{sec:mnar}


Note that knowing the parents of a mechanism variable $\R$ is effectively equivalent, for the purposes of informed deletion, to knowing the Markov blanket of the variables in $\R$ \citep{pearl1987evidential}, which can be learned from data \citep{tsamardinos2003algorithms,yaramakala2005speculative}.  With sufficient domain knowledge, an expert may be able to specify the parents of the mechanism variables. It suffices even to identify a subset of the observed variables that just \emph{contains} the Markov blanket; this knowledge can still be exploited to reduce the scope of the summation.  As we discuss next, having deeper knowledge of the nature of the missingness mechanisms, will enable us to obtain consistent estimators, even for datasets that are not MAR~(in some cases).

\subsection*{Empirical Evaluation of Informed Deletion}

Here, we evaluate the benefits of informed deletion. In addition to the MAR assumption, with this setting, we assume that we know the set of parents $\W_o$ of the missingness mechanism variables.

\begin{table*}[htbp]
  \caption{Alarm network with Informed MAR data}
  \label{tab:mar}
  \scriptsize
  \centering
\begin{tabular}{@{}r|r|rr|rr!{\vrule width 1pt}r|r|rr|rr@{}} \toprule

Size & F-MCAR & D-MAR & F-MAR & ID-MAR & IF-MAR & Size & F-MCAR & D-MAR & F-MAR & ID-MAR & IF-MAR\\

\hline\multicolumn{6}{c!{\vrule width 1pt}}{Kullback-Leibler Divergence} & \multicolumn{6}{c}{Test Set Log-Likelihood (Fully Observed)}\\\hline 
$10^{2}$ & 1.921 & 2.365 & 2.364 & 2.021 & 2.011      & $10^{2}$ & -11.67 & -12.13 & -12.13 & -11.77 & -11.76\\                    
$10^{3}$ & 0.380 & 0.454 & 0.452 & 0.399 & 0.375      & $10^{3}$ & -10.40 & -10.47 & -10.47 & -10.42 & -10.40\\                    
$10^{4}$ & 0.073 & 0.071 & 0.072 & 0.059 & 0.053      & $10^{4}$ & -10.04 & -10.04 & -10.04 & -10.02 & -10.02\\                    
$10^{5}$ & 0.041 & 0.021 & 0.022 & 0.011 & 0.010      & $10^{5}$ & -10.00 & -9.98 & -9.98 & -9.97 & -9.97\\                        
$10^{6}$ & 0.040 & 0.006 & 0.008 & 0.001 & 0.001      & $10^{6}$ & -10.00 & -9.97 & -9.97 & -9.96 & -9.96\\                        
\bottomrule

\end{tabular}
\end{table*}

To generate data for such a mechanism, we select a random set of $s$ variables to form $\W_o$. We further employ the sampling algorithm previously used for MAR data, but now insist that the parents of $\R$ variables come from $\W_o$.
Table~\ref{tab:mar} shows likelihoods and KLDs on the Alarm network, for $s=3$, and other settings as in the MAR experiments. Informed D-MAR (ID-MAR) and F-MAR (IF-MAR) consistently outperform their non-informed counterparts.

\subsection*{Missing Not at Random (MNAR) Data}

A missing data problem that is neither MCAR nor MAR is classified as
\emph{Missing Not at Random} (MNAR).  Here, the parameters of
a Bayesian network may not even be identifiable.  Further, maximum
likelihood estimation is in general not consistent, so the EM
algorithm and gradient methods are expected to yield biased estimates.
However, if one knows the interactions of the mechanisms that dictate
the missingness of a dataset (in the form of a missingness graph),
then it becomes possible again to obtain consistent estimates, at
least in some cases \citep{mohan2013missing}.

\begin{figure}[htbp]
  \centering
  \includegraphics[width=.50\linewidth,clip=true,angle=0]{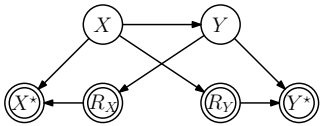}
  \caption{An MNAR missingness graph.}
  \label{fig:mnar}
\end{figure}

For example, consider
the missingness graph of Figure~\ref{fig:mnar},
which is an MNAR problem, where both variables \(X\) and \(Y\) are
partially observed, and the missingness of each variable depends on
the value of the other.  In this case, it is still possible to obtain
consistent parameter estimates, as
\begin{align*}
&\Pr(X,Y) \\
& \quad  = \frac{\Pr(\eql(R_X,\obs),\eql(R_Y,\obs)) \Pr(X^\star,Y^\star|\eql(R_X,\obs),\eql(R_Y,\obs))}
         {\Pr(\eql(R_X,\obs)|Y^\star,\eql(R_Y,\obs))
          \Pr(\eql(R_Y,\obs)|X^\star,\eql(R_X,\obs))}
\end{align*}
For a derivation, see \citet{mohan2013missing}.  Such derivations for
recovering queries under MNAR are extremely sensitive to the structure
of the missingness graph.  Indeed, the class of missingness graphs
that admit consistent estimation has not yet been fully characterized.
We view, as interesting future work, the identification of missingness
graph structures that guarantee consistent estimators (beyond MCAR and
MAR), under minimal assumptions (such as the ones we exploited for
informed deletion).

\shrink{
\begin{wraptable}{htb}{.55\linewidth}
  \caption{Alarm network with Informed MAR data}
  \label{tab:mar}
  \scriptsize
  \centering
\begin{tabular}{@{}r|r|rr|rr@{}} \toprule

Size & F-MCAR & D-MAR & F-MAR & ID-MAR & IF-MAR\\

\hline\multicolumn{6}{c}{Kullback-Leibler Divergence}\\\hline
$10^{2}$ & 1.921 & 2.365 & 2.364 & 2.021 & 2.011\\
$10^{3}$ & 0.380 & 0.454 & 0.452 & 0.399 & 0.375\\
$10^{4}$ & 0.073 & 0.071 & 0.072 & 0.059 & 0.053\\
$10^{5}$ & 0.041 & 0.021 & 0.022 & 0.011 & 0.010\\
$10^{6}$ & 0.040 & 0.006 & 0.008 & 0.001 & 0.001\\

\hline\multicolumn{6}{c}{Test Set Log-Likelihood (Fully Observed)}\\\hline
$10^{2}$ & -11.67 & -12.13 & -12.13 & -11.77 & -11.76\\
$10^{3}$ & -10.40 & -10.47 & -10.47 & -10.42 & -10.40\\
$10^{4}$ & -10.04 & -10.04 & -10.04 & -10.02 & -10.02\\
$10^{5}$ & -10.00 & -9.98 & -9.98 & -9.97 & -9.97\\
$10^{6}$ & -10.00 & -9.97 & -9.97 & -9.96 & -9.96\\

\bottomrule

\end{tabular}
\end{wraptable}
}

\shrink{
The general procedure for recovering queries under MNAR is extremely
sensitive to the structure of the missingness graph. In this paper, we
consider a subset of MNAR problems in which all partially-observed
variables can be separated from their respective missingness
mechanisms by variables that are either fully-observed or
partially-observed, such that their respective missingness mechanisms
are completely random. Formally, $\X_m$ can be partitioned into
$\X^1_m$ and $\X^2_m$ such that $\X^1_m \indep \R_{X^1_m}| \X^2_m$ and
$(\X^2_m , \X_o) \indep \R_{X^2_m} $.

**TODO**
}

\section{Extended Empirical Evaluation: MCAR} \label{sec:exp:mcar}

In this Appendix, we expand on the empirical results of Section~\ref{sec:experiments}.
Here, we investigate learning from \emph{MCAR data}, by generating MCAR datasets of increasing size, and evaluating the quality of the learned parameters for each algorithm.
\begin{table}[htb]
  \caption{Alarm network with MCAR data}
  \label{tab:mcar}
\scriptsize
  \centering
  \begin{tabular}{@{}r|rr|rr|rr@{}} \toprule                                                                        
                                                                                                                  
Size & EM-1-JT & EM-10-JT & D-MCAR & F-MCAR & D-MAR & F-MAR\\                                                     
\hline\multicolumn{7}{c}{Runtime [s]}\\\hline                                                                    
$10^{2}$ & 2 & 6 & 0 & 0 & 0 & 0\\                                                                         
$10^{3}$ & 6 & 50 & 0 & 0 & 0 & 0\\                                                                      
$10^{4}$ & 69 & - & 0 & 1 & 0 & 1\\                                                                
$10^{5}$ & - & - & 1 & 9 & 4 & 13\\
$10^{6}$ & - & - & 11 & 92 & 29 & 124\\

\hline\multicolumn{7}{c}{Test Set Log-Likelihood}\\\hline
$10^{2}$ & -12.18 & -12.18 & -12.85 & -12.33 & -12.82 & -12.32\\
$10^{3}$ & -10.41 & -10.41 & -10.73 & -10.55 & -10.69 & -10.55\\
$10^{4}$ & -10.00 & - & -10.07 & -10.04 & -10.07 & -10.05\\
$10^{5}$ & - & - & -9.98 & -9.98 & -9.99 & -9.98\\
$10^{6}$ & - & - & -9.96 & -9.96 & -9.97 & -9.97\\

\hline\multicolumn{7}{c}{Kullback-Leibler Divergence}\\\hline
$10^{2}$ & 2.381 & 2.381 & 3.037 & 2.525 & 3.010 & 2.515\\
$10^{3}$ & 0.365 & 0.365 & 0.688 & 0.502 & 0.659 & 0.502\\
$10^{4}$ & 0.046 & - & 0.113 & 0.084 & 0.121 & 0.093\\
$10^{5}$ & - & - & 0.016 & 0.013 & 0.024 & 0.021\\
$10^{6}$ & - & - & 0.002 & 0.002 & 0.006 & 0.008\\

\bottomrule

\end{tabular}
%
%
\end{table}

Table~\ref{tab:mcar} shows results for the ``Alarm'' Bayesian network.
Each training set is simulated from the original Bayesian network, selecting 30\% of the variables to be partially observed, and removing 70\% of their values completely at random.
All reported numbers are averaged over 32 repetitions with different learning problems. When no number is reported, a 5 minute time limit was exceeded.

We first note that there is no advantage in running EM with restarts: EM-1-JT and EM-10-JT learn almost identical models. This indicates that the likelihood landscape for MCAR data has few local optima, and is easy to optimize.
Direct and factored deletion are orders of magnitude faster than EM, which needs to repeatedly run inference for every instance in the dataset.
Even though EM outperforms F-MCAR in terms of KLD and likelihood, the difference is negligible, in the sense that only a small difference in the amount of available data makes F-MCAR outperform EM.
F-MCAR is slower than D-MCAR, because it requires estimating more probabilities (one for each lattice edge). F-MCAR does learn better models, because it can use a larger portion of the available data.
Finally, D-MAR performs worse than F-MCAR and D-MCAR, as it is operating on the weaker MAR assumption.
All learners are consistent, as can be seen from the KLD converging to zero.

\begin{figure}[ht]
  \centering
  \subfigure[KL Divergence vs.\ Dataset Size]{%
    \includegraphics[height=\plotheight]{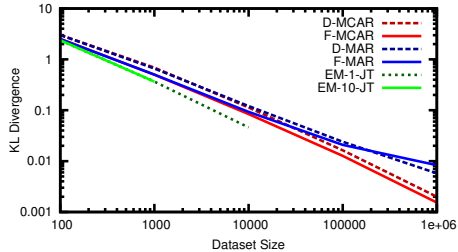}
    \label{fig:mcar:a}
  } \\
  \subfigure[KL Divergence vs.\ Time]{%
    \includegraphics[height=\plotheight]{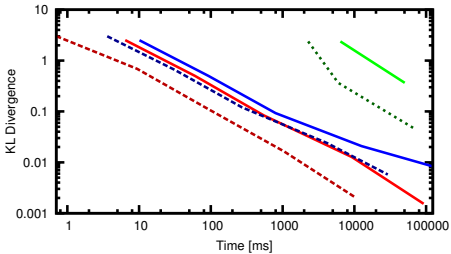}~~
    \label{fig:mcar:b}
  }
  \caption{Learning the ``Alarm'' network from MCAR data, as show in Table~\ref{tab:mcar}.}
  \label{fig:mcar}
\end{figure}

To illustrate the trade-off between data and computational resources, Figure~\ref{fig:mcar} shows the KLDs from Table~\ref{tab:mcar} as a function of dataset size and time.
When data is limited, and computation power is abundant, it is clear that EM is the algorithm of choice, even though the differences are small.
When computation power is limited (e.g., when the Bayesian network is highly intractable), and data is abundant (e.g., the online learning or big data setting), the differences are marked. EM is several orders of magnitudes slower than D-MCAR at learning a model of similar quality.
F-MCAR may provide a good trade-off.

\section{Extended Empirical Evaluation: MAR} \label{sec:extended}

In this Appendix, we expand on the empirical results of Section~\ref{sec:experiments} w.r.t.\ learning from MAR data.  Here, we provide additional empirical results on standard real-world networks where inference is challenging, as originally highlighted in Table~\ref{table:bp}.

\begin{table*}[htbp]
\scriptsize
  \centering
  \caption{Log-likelihoods of large networks learned from MAR data (1 min.~time limit, 1st setting).\label{tab:large:a}}
  \begin{tabular}{@{}r!{\vrule width 1pt}r|rr|rr|rr!{\vrule width 1pt}r|rr|rr|rr@{}} \toprule
Size &  & EM-JT & EM-BP & D-MCAR & F-MCAR & D-MAR & F-MAR &  & EM-JT & EM-BP & D-MCAR & F-MCAR & D-MAR & F-MAR\\
\hline
$10^{2}$ & \multirow{4}{*}{\rotatebox[origin=c]{90}{Grid 90-20-1}} & - & -62.38 & -64.15 & -50.78 & -63.51 & \textbf{-50.24} & \multirow{4}{*}{\rotatebox[origin=c]{90}{Water}} & - & -19.50 & -20.51 & -19.37 & -20.41 & \textbf{-19.35} \\
$10^{3}$ & & - & -79.75 & -38.96 & -32.77 & -38.26 & \textbf{-32.44} & & - & -16.11 & -16.26 & -15.27 & -16.09 & \textbf{-15.23} \\
$10^{4}$ & & - & - & -30.65 & -28.61 & -30.05 & \textbf{-28.34} & & - & - & -15.03 & -14.22 & -14.86 & \textbf{-14.14} \\
$10^{5}$ & & - & - & - & - & - & - & & - & - & \textbf{-14.30} & - & - & - \\
\hline
$10^{2}$ & \multirow{3}{*}{\rotatebox[origin=c]{90}{Munin 1}} & - & -98.95 & -103.59 & -98.68 & -103.54 & \textbf{-98.49} & \multirow{3}{*}{\rotatebox[origin=c]{90}{Barley}} & - & \textbf{-85.33} & -85.84 & -85.68 & -86.13 & -85.75 \\
$10^{3}$ & & - & -79.83 & -70.49 & -67.27 & -69.78 & \textbf{-66.97} & & - & - & -67.70 & -67.18 & -67.67 & \textbf{-67.13} \\
$10^{4}$ & & - & - & -59.25 & \textbf{-57.11} & - & - & & - & - & \textbf{-54.93} & - & - & - \\
\bottomrule
\end{tabular}

\end{table*} 

\begin{table*}[htbp]
\scriptsize
  \centering
  \caption{Log-likelihoods of large networks learned from MAR data (5 min.~time limit, 1st setting).}
  \begin{tabular}{@{}r!{\vrule width 1pt}r|rr|rr|rr!{\vrule width 1pt}r|rr|rr|rr@{}} \toprule
Size &  & EM-JT & EM-BP & D-MCAR & F-MCAR & D-MAR & F-MAR &  & EM-JT & EM-BP & D-MCAR & F-MCAR & D-MAR & F-MAR\\
\hline
$10^{2}$ & \multirow{5}{*}{\rotatebox[origin=c]{90}{Grid 90-20-1}} & - & -56.23 & -63.34 & -50.55 & -62.38 & \textbf{-50.06} & \multirow{5}{*}{\rotatebox[origin=c]{90}{Water}} & -18.84 & \textbf{-18.06} & -21.23 & -19.61 & -21.07 & -19.57 \\
$10^{3}$ & & - & -55.04 & -39.89 & -33.34 & -39.09 & \textbf{-33.01} & & - & \textbf{-14.99} & -16.47 & -15.33 & -16.24 & -15.26 \\
$10^{4}$ & & - & -98.20 & -30.46 & -27.26 & -29.73 & \textbf{-26.98} & & - & -17.39 & -15.59 & -14.52 & -15.26 & \textbf{-14.43} \\
$10^{5}$ & & - & - & -28.63 & \textbf{-26.06} & -27.89 & - & & - & - & \textbf{-15.22} & - & - & - \\
$10^{6}$ & & - & - & - & - & - & - & & - & - & \textbf{-15.09} & - & - & - \\
\hline
$10^{2}$ & \multirow{4}{*}{\rotatebox[origin=c]{90}{Munin 1}} & - & \textbf{-96.51} & -102.51 & -98.21 & -102.40 & -97.95 & \multirow{4}{*}{\rotatebox[origin=c]{90}{Barley}} & - & \textbf{-85.59} & -85.70 & -85.60 & -85.99 & -85.66 \\
$10^{3}$ & & - & -68.04 & -67.82 & -65.49 & -67.21 & \textbf{-65.22} & & - & -67.07 & -67.58 & -66.97 & -67.53 & \textbf{-66.91} \\
$10^{4}$ & & - & -95.01 & -57.68 & -56.00 & -57.05 & \textbf{-55.79} & & - & - & -55.04 & \textbf{-54.33} & -54.78 & - \\
$10^{5}$ & & - & - & \textbf{-54.30} & - & - & - & & - & - & - & - & - & - \\
\bottomrule
\end{tabular}

\end{table*} 

\begin{table*}[htbp]
\scriptsize
  \centering
  \caption{Log-likelihoods of large networks learned from MAR data (25 min.~time limit, 1st setting).}
  \begin{tabular}{@{}r!{\vrule width 1pt}r|rr|rr|rr!{\vrule width 1pt}r|rr|rr|rr@{}} \toprule
Size &  & EM-JT & EM-BP & D-MCAR & F-MCAR & D-MAR & F-MAR &  & EM-JT & EM-BP & D-MCAR & F-MCAR & D-MAR & F-MAR\\
\hline
$10^{2}$ & \multirow{6}{*}{\rotatebox[origin=c]{90}{Grid 90-20-1}} & - & \textbf{-47.66} & -59.84 & -48.34 & -59.39 & -47.88 & \multirow{6}{*}{\rotatebox[origin=c]{90}{Water}} & -21.30 & \textbf{-18.66} & -21.58 & -19.87 & -21.36 & -19.83 \\
$10^{3}$ & & - & -46.53 & -37.29 & -31.60 & -36.76 & \textbf{-31.28} & & -17.67 & -17.10 & -18.64 & -15.95 & -18.27 & \textbf{-15.86} \\
$10^{4}$ & & - & -62.98 & -28.74 & -26.71 & -28.26 & \textbf{-26.45} & & - & -14.83 & -16.71 & -14.58 & -16.30 & \textbf{-14.44} \\
$10^{5}$ & & - & - & -25.88 & -24.97 & -25.43 & \textbf{-24.75} & & - & -18.78 & -16.31 & -14.38 & -15.62 & \textbf{-14.08} \\
$10^{6}$ & & - & - & -25.27 & - & \textbf{-24.78} & - & & - & - & \textbf{-15.25} & - & - & - \\
$10^{7}$ & & - & - & - & - & - & - & & - & - & \textbf{-15.13} & - & - & - \\
\hline
$10^{2}$ & \multirow{4}{*}{\rotatebox[origin=c]{90}{Munin 1}} & - & \textbf{-90.79} & -98.57 & -94.50 & -98.48 & -94.28 & \multirow{4}{*}{\rotatebox[origin=c]{90}{Barley}} & \textbf{-85.11} & -85.53 & -86.00 & -85.74 & -86.24 & -85.80 \\
$10^{3}$ & & - & \textbf{-60.71} & -66.06 & -63.95 & -65.45 & -63.67 & & - & \textbf{-65.96} & -67.88 & -67.23 & -67.79 & -67.15 \\
$10^{4}$ & & - & -60.35 & -56.57 & -55.38 & -55.95 & \textbf{-55.16} & & - & -57.21 & -55.34 & -54.56 & -55.05 & \textbf{-54.43} \\
$10^{5}$ & & - & - & -54.29 & \textbf{-53.38} & -53.67 & - & & - & - & \textbf{-51.09} & - & - & - \\
\bottomrule
\end{tabular}

\end{table*} 

\begin{table*}[htbp]
\scriptsize
  \centering
  \caption{Log-likelihoods of large networks learned from MAR data (1 min.~time limit, 2nd setting).}
  \begin{tabular}{@{}r!{\vrule width 1pt}r|rr|rr|rr!{\vrule width 1pt}r|rr|rr|rr@{}} \toprule
Size &  & EM-JT & EM-BP & D-MCAR & F-MCAR & D-MAR & F-MAR &  & EM-JT & EM-BP & D-MCAR & F-MCAR & D-MAR & F-MAR\\
\hline
$10^{2}$ & \multirow{4}{*}{\rotatebox[origin=c]{90}{Grid 90-20-1}} & - & -62.25 & -80.10 & -56.59 & -79.93 & \textbf{-56.07} & \multirow{4}{*}{\rotatebox[origin=c]{90}{Water}} & - & \textbf{-20.15} & -26.40 & -22.85 & -26.24 & -22.88 \\
$10^{3}$ & & - & -129.38 & -38.74 & -29.88 & -38.51 & \textbf{-29.70} & & - & -17.76 & -20.45 & -17.80 & -20.32 & \textbf{-17.64} \\
$10^{4}$ & & - & - & -27.83 & -24.30 & -27.25 & \textbf{-23.97} & & - & - & -17.59 & -15.40 & -17.28 & \textbf{-15.29} \\
$10^{5}$ & & - & - & - & - & - & - & & - & - & \textbf{-15.38} & - & - & - \\
\hline
$10^{2}$ & \multirow{3}{*}{\rotatebox[origin=c]{90}{Munin 1}} & - & \textbf{-99.49} & -111.95 & -104.07 & -111.72 & -103.10 & \multirow{3}{*}{\rotatebox[origin=c]{90}{Barley}} & - & -89.16 & -89.63 & -89.13 & -89.66 & \textbf{-88.99} \\
$10^{3}$ & & - & -99.56 & -70.32 & -66.08 & -69.76 & \textbf{-65.57} & & - & - & -71.76 & \textbf{-70.50} & -71.74 & - \\
$10^{4}$ & & - & - & -56.25 & \textbf{-54.36} & - & - & & - & - & \textbf{-56.59} & - & - & - \\
\bottomrule
\end{tabular}

\end{table*} 

\begin{table*}[htbp]
\scriptsize
  \centering
  \caption{Log-likelihoods of large networks learned from MAR data (5 min.~time limit, 2nd setting).\label{tab:large:z}}
  \begin{tabular}{@{}r!{\vrule width 1pt}r|rr|rr|rr!{\vrule width 1pt}r|rr|rr|rr@{}} \toprule
Size &  & EM-JT & EM-BP & D-MCAR & F-MCAR & D-MAR & F-MAR &  & EM-JT & EM-BP & D-MCAR & F-MCAR & D-MAR & F-MAR\\
\hline
$10^{2}$ & \multirow{4}{*}{\rotatebox[origin=c]{90}{Grid 90-20-1}} & - & -57.14 & -80.92 & -57.01 & -80.80 & \textbf{-56.53} & \multirow{4}{*}{\rotatebox[origin=c]{90}{Water}} & -19.10 & \textbf{-18.76} & -25.31 & -21.76 & -25.29 & -21.81 \\
$10^{3}$ & & - & -65.41 & -38.54 & -30.07 & -38.27 & \textbf{-29.86} & & - & \textbf{-14.73} & -19.13 & -16.45 & -18.93 & -16.36 \\
$10^{4}$ & & - & - & -25.95 & -23.30 & -25.36 & \textbf{-22.88} & & - & -20.70 & -16.66 & -14.90 & -16.33 & \textbf{-14.67} \\
$10^{5}$ & & - & - & -22.74 & -22.01 & \textbf{-21.60} & - & & - & - & -15.49 & - & \textbf{-14.90} & - \\
\hline
$10^{2}$ & \multirow{4}{*}{\rotatebox[origin=c]{90}{Munin 1}} & - & \textbf{-103.72} & -115.50 & -105.81 & -115.41 & -104.87 & \multirow{4}{*}{\rotatebox[origin=c]{90}{Barley}} & - & -89.22 & -89.54 & -89.26 & -89.60 & \textbf{-89.14} \\
$10^{3}$ & & - & -69.03 & -71.01 & -65.91 & -70.61 & \textbf{-65.51} & & - & -74.26 & -71.67 & -70.46 & -71.68 & \textbf{-70.18} \\
$10^{4}$ & & - & -157.23 & -56.07 & \textbf{-54.24} & -55.46 & - & & - & - & -56.44 & \textbf{-55.12} & -56.40 & - \\
$10^{5}$ & & - & - & \textbf{-52.00} & - & - & - & & - & - & - & - & - & - \\
\bottomrule
\end{tabular}

\end{table*} 


We consider two settings of generating MAR data, as in Section~\ref{sec:experiments}. In the \emph{first setting}, the missing data mechanisms were generated with $m=0.3$, $p=2$, and a Beta distribution with shape parameters $1.0$ and $0.5$. 
In the second setting, we have $m=0.9$, $p=2$, and a Beta distribution with shape parameters $0.5$.
We consider three time limits, of 1 minute, 5 minutes, and 25 minutes. For all combinations of these setting, test set log-likelihoods are shown in Table~\ref{table:bp}, and in Tables~\ref{tab:large:a} to~\ref{tab:large:z}.  

We repeat the observations from the main paper (cf. Section~\ref{sec:experiments}).
The EM-JT learner, which performs exact inference, does not scale well to these networks. This problem is mitigated by EM-BP, which performs  \emph{approximate} inference, yet we find that it also has difficulties scaling (dashed entries indicate that EM-JT and EM-BP did not finish 1 iteration of EM).  
In contrast, F-MAR, and particularly D-MAR, can scale to much larger datasets.  
As for accuracy, the F-MAR method typically obtains the best likelihoods (in bold) for larger datasets, although EM-BP can perform better on small datasets.  We further evaluated D-MCAR and F-MCAR, although they are not in general consistent for MAR data, and find that they scale even further, and can also produce relatively good estimates (in terms of likelihood).

\section{Data Exploitation by Closed-Form Parameter Learners: Example} \label{app:example} \label{sec:fmcarexample}
This appendix demonstrates with an example how each learning 
algorithm exploits varied subsets of data to estimate 
marginal probability distributions,  given the manifest (or data)
distribution 
in Table \ref{tab:1} which  consists of four variables, $\{X,Y,Z,W\}$ such that $\{X,Y \} \in \X_m$ and $\{Z,W\}\in \X_o$.

\begin{table*}[htbp]
  \caption{Manifest (Data) Distribution with $\{X,Y\} \in \X_m$ and $\{Z,W\} \in \X_o$. }
  \label{tab:1}
  \footnotesize
  \centering
  \begin{tabular}{|c|c|c|c|c|c|c|} \hline
$\#$ & $X$ & $Y$ & $W$ & $Z$ & $R_x$ & $R_y$ \\ \hline
$1$ & $0$ & $0$ & $0$ & $0$ & $\obs$ & $\obs$ \\
$2$ & $0$ & $0$ & $0$ & $1$ & $\obs$ & $\obs$ \\
$3$ & $0$ & $0$ & $1$ & $0$ & $\obs$ & $\obs$ \\
$4$ & $0$ & $0$ & $1$ & $1$ & $\obs$ & $\obs$ \\
$5$ & $0$ & $1$ & $0$ & $0$ & $\obs$ & $\obs$ \\
$6$ & $0$ & $1$ & $0$ & $1$ & $\obs$ & $\obs$ \\
$7$ & $0$ & $1$ & $1$ & $0$ & $\obs$ & $\obs$ \\
$8$ & $0$ & $1$ & $1$ & $1$ & $\obs$ & $\obs$ \\
$9$ & $1$ & $0$ & $0$ & $0$ & $\obs$ & $\obs$ \\
$10$ & $1$ & $0$ & $0$ & $1$ & $\obs$ & $\obs$ \\
$11$ & $1$ & $0$ & $1$ & $0$ & $\obs$ & $\obs$ \\
$12$ & $1$ & $0$ & $1$ & $1$ & $\obs$ & $\obs$ \\
$13$ & $1$ & $1$ & $0$ & $0$ & $\obs$ & $\obs$ \\
$14$ & $1$ & $1$ & $0$ & $1$ & $\obs$ & $\obs$ \\
$15$ & $1$ & $1$ & $1$ & $0$ & $\obs$ & $\obs$ \\ 
$16$ & $1$ & $1$ & $1$ & $1$ & $\obs$ & $\obs$ \\ \hline
$17$ & $0$ & $?$ & $0$ & $0$ & $\obs$ & $\unobs$ \\
$18$ & $0$ & $?$ & $0$ & $1$ & $\obs$ & $\unobs$ \\\hline
\end{tabular}
\qquad\qquad
\begin{tabular}{|c|c|c|c|c|c|c|} \hline
$\#$ & $X$ & $Y$ & $W$ & $Z$ & $R_x$ & $R_y$ \\ \hline
$19$ & $0$ & $?$ & $1$ & $0$ & $\obs$ & $\unobs$ \\
$20$ & $0$ & $?$ & $1$ & $1$ & $\obs$ & $\unobs$ \\ \hline
$21$ & $1$ & $?$ & $0$ & $0$ & $\obs$ & $\unobs$ \\
$22$ & $1$ & $?$ & $0$ & $1$ & $\obs$ & $\unobs$ \\
$23$ & $1$ & $?$ & $1$ & $0$ & $\obs$ & $\unobs$ \\
$24$ & $1$ & $?$ & $1$ & $1$ & $\obs$ & $\unobs$ \\ \hline
$25$ & $?$ & $0$ & $0$ & $0$ & $\unobs$ & $\obs$ \\
$26$ & $?$ & $0$ & $0$ & $1$ & $\unobs$ & $\obs$ \\
$27$ & $?$ & $0$ & $1$ & $0$ & $\unobs$ & $\obs$ \\
$28$ & $?$ & $0$ & $1$ & $1$ & $\unobs$ & $\obs$ \\ \hline
$29$ & $?$ & $1$ & $0$ & $0$ & $\unobs$ & $\obs$ \\
$30$ & $?$ & $1$ & $0$ & $1$ & $\unobs$ & $\obs$ \\
$31$ & $?$ & $1$ & $1$ & $0$ & $\unobs$ & $\obs$ \\
$32$ & $?$ & $1$ & $1$ & $1$ & $\unobs$ & $\obs$ \\ \hline
$33$ & $?$ & $?$ & $0$ & $0$ & $\unobs$ & $\unobs$ \\
$34$ & $?$ & $?$ & $0$ & $1$ & $\unobs$ & $\unobs$ \\
$35$ & $?$ & $?$ & $1$ & $0$ & $\unobs$ & $\unobs$ \\
$36$ & $?$ & $?$ & $1$ & $1$ & $\unobs$ & $\unobs$ \\ \hline
\end{tabular}
\end{table*}
\begin{table*}[htbp]  \caption{Enumeration of sample $\#$ used for computing $\Pr(x,w)$ by listwise deletion, direct deletion and factored deletion algorithms under MCAR assumptions.}
  \label{tab:2}
  \footnotesize
  \centering
\begin{tabular}{|l|l|} \hline
Algorithm & Estimand and Sample $\#$  \\ \hline
Listwise &  \parbox[t]{3in}{$\Pr(xw)= \Pr(xw|R_x=\obs,R_y=\obs)$ \par\quad 11,12,15,16 \strut}  \\ \hline
Direct &  \parbox[t]{3in}{$\Pr(xw)= \Pr(xw|R_x=\obs)$ \par\quad 11,12,15,16,23,24 \strut}     \\ \hline
Factored & \parbox[t]{3in}{$\Pr(xw)= \Pr(x|w,R_x=\obs)\Pr(w)$ \par \quad 3,4,7,8,11,12,15,16,19,20,23,24,27,28,31,32,35,36 \par $\Pr(xw)= \Pr(w|x,R_x=\obs)\Pr(x|R_x=\obs)$ \par\quad 9,10,11,12,13,14,15,16,21,22,23,24 \strut}  \\ \hline

\end{tabular}

\end{table*}
\begin{table*}[htb]
  \caption{Enumeration of sample $\#$ used for computing $\Pr(x,y)$ by direct deletion, factored deletion and informed deletion algorithms under MAR assumption.}
  \label{tab:3}
  \footnotesize
  \centering
\begin{tabular}{|l|l|} \hline
Algorithm & Estimand and Sample $\#$ \\ \hline
    Direct & \parbox[t]{3in}{$\Pr(xy)= \sum_{z,w} \Pr(xy|w,z,R_x=\obs,R_y=\obs)\Pr(zw)$ \par\quad 13,14,15,16 for $\Pr(xy|w,z,R_x=\obs,R_y=\obs)$\par \quad all tuples: [1,36] for $\Pr(z,w)$ \strut}   \\ \hline
        Factored & \parbox[t]{3.5in}{$\Pr(xy)= \sum_{z,w} \Pr(x|w,z,y,R_x=\obs,R_y=\obs)$ \par \qquad\qquad\qquad\quad $\Pr(y|z,w,R_y=\obs)\Pr(zw)$ \par\quad13,14,15,16 for $\Pr(x|y,w,z,R_x=\obs,R_y=\obs)$
        \par\quad 5,6,7,8,13,14,15,16,29,30,31,32 for $\Pr(y|w,z,R_y=\obs)$ \par\quad all tuples: [1,36] for $\Pr(z,w)$\\ \par $\Pr(xy)= \sum_{z,w} \Pr(y|x,w,z,R_x=\obs,R_y=\obs)$ \par \qquad\qquad\qquad\quad $\Pr(x|z,w,R_x=\obs)\Pr(zw)$ \par\quad13,14,15,16 for $\Pr(y|x,w,z,R_x=\obs,R_y=\obs)$
        \par\quad 9,10,11,12,13,14,15,16,21,22,23,24 for $\Pr(x|w,z,R_x=\obs)$ \par\quad all tuples: [1,36] for $\Pr(z,w)$ \strut}    \\ \hline
        \parbox[t]{1.5in}{Informed (direct)  \par \raisebox{-.9\height}{\includegraphics[scale=0.5]{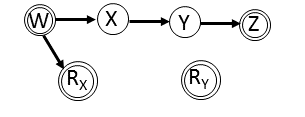}} \strut}  & \parbox[t]{3in}{ $\Pr(xy)=\sum_w \Pr(xy|w,R_x=\obs,R_y=\obs)\Pr(w)$ \par \quad 13,14,15,16 for $\Pr(xy|w,R_x=\obs,R_y=\obs)$ \par \quad all tuples: [1,36] for $\Pr(w)$ \strut} \\ 
\hline
\end{tabular}
\end{table*}

We will begin by examining the data usage by deletion algorithms while estimating $\Pr(x,w)$  under the MCAR assumption.
All three deletion algorithms, namely listwise deletion, direct deletion and factored deletion guarantee consistent estimates when data are MCAR. Among these
 algorithms, listwise deletion utilizes the least 
amount of data (4 distinct tuples out of 36 available tuples, as shown in table \ref{tab:2}) to compute $\Pr(xw)$ where as factored 
deletion employs two thirds of the tuples (24 distinct tuples out of 36 available tuples as shown in table \ref{tab:2}) for estimating $\Pr(xw)$. 
%
%

Under MAR, no guarantees are available for listwise deletion. However the three algorithms, namely direct deletion, factored deletion and informed deletion,  guarantee consistent estimates. While estimating $\Pr(x,y)$, all the three algorithms utilize every tuple in the manifest distribution at least once (see table \ref{tab:3}).  Compared to direct deletion algorithm, the factored deletion algorithm utilizes more data while computing $\Pr(x,y)$  since it has multiple factorizations with more than two factors in each of them; this allows more data to be used while computing each factor (see table \ref{tab:2}).  In contrast to both direct and factored deletion, the informed deletion algorithm yields an estimand that involves factors with fewer elements in them ($\Pr(w)$ vs.\ $\Pr(zw)$) and hence can be computed using more data ($\Pr(w=0)$ uses 18 tuples compared to $\Pr(z=0,w=0)$ that uses 9 tuples).

Precise information regarding the missingness process is required for estimation when dataset falls under the MNAR category. In particular, only algorithms that consult the missingness graph can answer questions about estimability of queries.

\end{document}